\documentclass[sigconf]{acmart}
\AtBeginDocument{%
  }

\usepackage{subcaption}
\usepackage{amsmath}
\usepackage{amsfonts}
\usepackage{algorithmic}
\usepackage[ruled,vlined]{algorithm2e}
\usepackage{bm}
\copyrightyear{2026}
\acmYear{2026}
\setcopyright{cc}
\setcctype{by-nc-nd}
\acmConference[KDD '26]{Proceedings of the 32nd ACM SIGKDD Conference on Knowledge Discovery and Data Mining V.1}{August 09--13, 2026}{Jeju Island, Republic of Korea}
\acmBooktitle{Proceedings of the 32nd ACM SIGKDD Conference on Knowledge Discovery and Data Mining V.1 (KDD '26), August 09--13, 2026, Jeju Island, Republic of Korea}
\acmDOI{10.1145/3770854.3780325}
\acmISBN{979-8-4007-2258-5/2026/08}

\begin{document}

%%
%% The "title" command has an optional parameter,
%% allowing the author to define a "short title" to be used in page headers.
\title{Holistic Data Scheduler for LLM Pre-training via Multi-Objective Reinforcement Learning}

%%
%% The "author" command and its associated commands are used to define
%% the authors and their affiliations.
%% Of note is the shared affiliation of the first two authors, and the
%% "authornote" and "authornotemark" commands
%% used to denote shared contribution to the research.

\author{Chenhao Dang}
\email{dangchenhao@std.uestc.edu.cn}
\orcid{0009-0001-8608-4078}
\affiliation{%
  \institution{China Electronics Technology Group Corporation 15th Research Institute}
  \city{Beijing}
  \country{China}
}

\author{Jing Ma}
\authornote{Corresponding author of this paper.}
\email{majingmady@ruc.edu.cn}
\orcid{0000-0002-0076-8189}
\affiliation{%
  \institution{Renmin University of China}
  \department{BRAIN}
  \city{Beijing}
  \country{China}
}

\author{Mingjie Liao}
\email{liaomingjie.lmj@alibaba-inc.com}
\orcid{0000-0002-7096-0766}
\affiliation{%
  \institution{Alibaba Group}
  \city{Beijing}
  \country{China}
}

%%
%% By default, the full list of authors will be used in the page
%% headers. Often, this list is too long, and will overlap
%% other information printed in the page headers. This command allows
%% the author to define a more concise list
%% of authors' names for this purpose.

%%
%% The abstract is a short summary of the work to be presented in the
%% article.
\begin{abstract}
  The composition of training data, governed by the diversity of sources and their mixing strategy, is a cornerstone of Large Language Model (LLM) pre-training. Online Data Mixing (ODM), the technique of adaptively adjusting data mixtures during training, has emerged as a promising direction to improve efficiency. However, existing methods are constrained by their reliance on a singular optimization perspective, which fundamentally overlooks the need for complex LLM pre-training to consider the dynamic data composition from multiple dimensions. To overcome this limitation, we introduce the Holistic Data Scheduler (HDS), a novel online data mixing framework. HDS formulates the data scheduling challenge as a reinforcement learning problem in a continuous control space and leverages the Soft Actor-Critic (SAC) algorithm for its stability and sample efficiency in exploring the high-dimensional policy space. At the core of HDS lies a novel multi-objective, holistic reward function that integrates three critical perspectives: a data-driven reward for quality, a loss-driven reward capturing inter-domain influence, and a model-driven reward based on weight norms. To validate our design and determine its optimal configuration, we conducted systematic experiments on LLMs of various sizes. On The Pile benchmark, HDS reaches the final validation perplexity of the next best method with 44\% fewer training iterations. Furthermore, it achieves a 7.2\% improvement on the MMLU 0-shot task along with consistent gains on other benchmarks, showcasing its ability to enhance both training efficiency and final model capability.
\end{abstract}

%%
%% The code below is generated by the tool at http://dl.acm.org/ccs.cfm.
%% Please copy and paste the code instead of the example below.
%%


%%
%% Keywords. The author(s) should pick words that accurately describe
%% the work being presented. Separate the keywords with commas.
%% A "teaser" image appears between the author and affiliation
%% information and the body of the document, and typically spans the
%% page.
\begin{teaserfigure}
  \begin{subfigure}[t]{0.42\textwidth}
        \includegraphics[width=\textwidth]{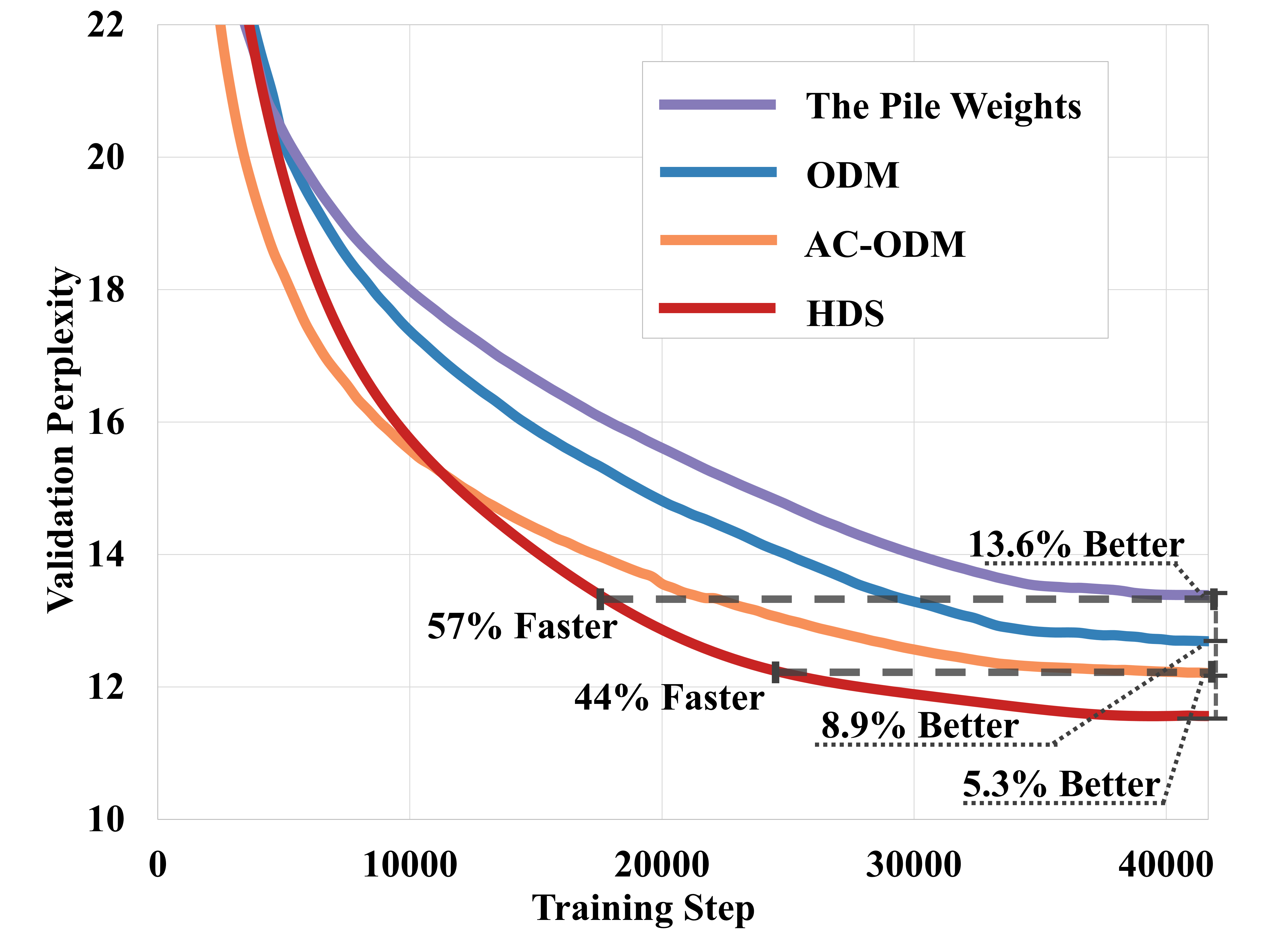}
        \caption{}
        \label{fig:f1a}
    \end{subfigure}
    \hfill
    \begin{subfigure}[t]{0.42\textwidth}
        \includegraphics[width=\textwidth]{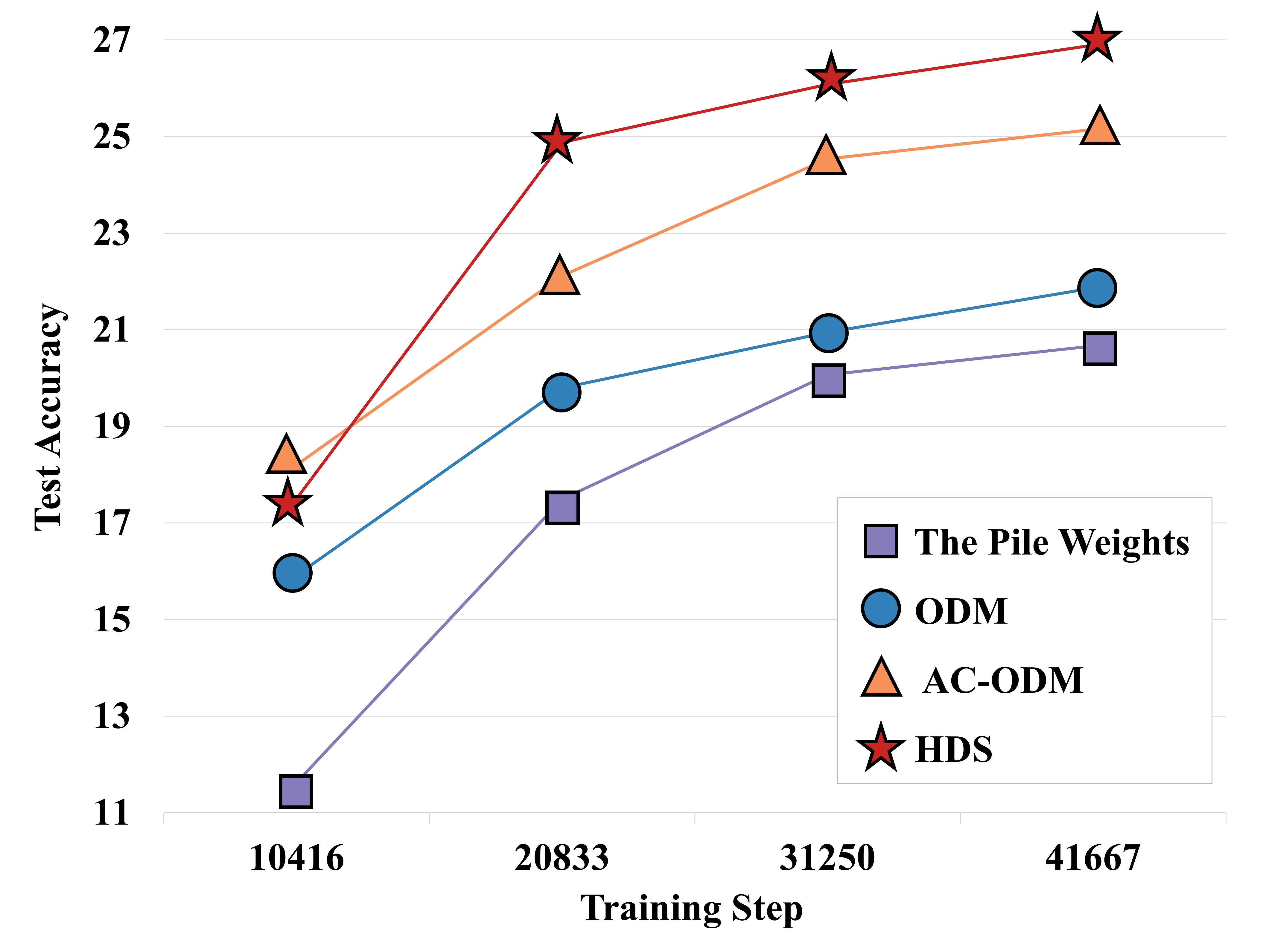}
        \caption{}
        \label{fig:f1b}
    \end{subfigure}
  \caption{Performance evaluation of the Holistic Data Scheduler (HDS). All results are from a Pythia-1B model trained for 50 billion tokens on The Pile dataset. (a) Unweighted average validation perplexity across the 22 domains of The Pile. (b) 0-shot accuracy on the MMLU benchmark.}
  \Description{Performance evaluation of the Holistic Data Scheduler (HDS).}
  \label{fig:f1}
\end{teaserfigure}

%%
%% This command processes the author and affiliation and title
%% information and builds the first part of the formatted document.
\maketitle

\newcommand\kddavailabilityurl{https://doi.org/10.5281/zenodo.18123749}
\ifdefempty{\kddavailabilityurl}{}{
\begingroup\small\noindent\raggedright\textbf{Resource Availability:}\\
% please change the following context to include multiple artifacts if necessary, including data, models, code, etc.
The source code of this paper has been made publicly available at \url{\kddavailabilityurl}.
\endgroup
}

\section{Introduction}

The paradigm for enhancing Large Language Models (LLMs) during pre-training is undergoing a fundamental shift from the compute-centric Scaling Laws towards Data-Centric AI ~\cite{matarazzo2025survey, zhao2023survey, peng2025unsupervised, du2022glam}. The quality, diversity, and mixing strategy of pre-training data are now recognized as decisive factors that not only shape the model's ultimate capabilities but also have profound economic and environmental implications ~\cite{ye2024data}. A meticulously designed data mixing strategy can significantly accelerate training convergence, thereby substantially reducing the immense financial costs and carbon footprint associated with pre-training ~\cite{ma2025actor, xie2023doremi}.

Research in data mixing strategies can be broadly categorized into static offline and dynamic online methods. Static approaches often leverage computationally inexpensive proxy models to discover an optimal mixing strategy ~\cite{albalak2024survey}. Prominent examples include DoReMi~\cite{xie2023doremi}, which trains a proxy model with Group Distributionally Robust Optimization (Group DRO) to enhance robustness against unknown downstream tasks, and DoGE~\cite{fan2024doge}, which formulates the problem as a bi-level optimization to explicitly maximize generalization on target domains. Similarly, RegMix~\cite{liu2024regmix} employs a regression model to fit the relationship between mixing ratios and model performance, thereby efficiently identifying an optimal recipe. However, a fundamental limitation of these static methods is their core assumption: that a single data recipe remains optimal throughout the entire training process. This paradigm fails to account for the inherently non-stationary nature of an LLM's learning dynamics.

To address this, online data mixing (ODM) techniques have emerged, which adapt domain weights at each training step based on the model's current state. The pioneering work, ODM~\cite{albalak2023efficient}, ingeniously models this as a Multi-Armed Bandit (MAB) problem~\cite{auer2002nonstochastic}. While effective, it uses the training loss on each domain's batch directly as a reward. This reward signal is myopic, as it evaluates each domain in isolation and fails to capture potential inter-domain synergies. To overcome this limitation, AC-ODM~\cite{ma2025actor} advanced the approach by employing the DDPG reinforcement learning algorithm~\cite{ma2025actor}, with its core innovation lying in a reward mechanism based on "gradient alignment." More recent works, such as AutoScale~\cite{kang2024autoscale} and Aioli~\cite{chenaioli}, have further refined the dynamic approach by integrating more comprehensive frameworks. AutoScale introduces a two-stage, scale-aware data composition framework that adapts to evolving training scales, offering a more scalable solution. Aioli, on the other hand, dynamically adjusts data proportions using an online method that more accurately estimates mixing law parameters, improving performance across multiple datasets. These innovations address the limitations of earlier approaches by considering the broader context of model performance, learning dynamics, and scale. Despite these advances, existing dynamic strategies still suffer from a narrow focus. They tend to optimize domain weights from a single perspective—such as data quality, training loss, or inter-domain influence—while neglecting the need to holistically consider the data composition from multiple dimensions, a crucial requirement for the complex LLM pre-training process.

To overcome this limitation, we propose the Holistic Data Scheduler (HDS), a novel online data mixing framework centered on a multi-objective reward system. HDS formulates the data mixing challenge as a reinforcement learning task in a continuous control domain, formalized as a Markov Decision Process (MDP). We employ the Soft Actor-Critic (SAC) algorithm~\cite{haarnoja2018soft}, renowned for its stability and sample efficiency in exploring high-dimensional policy spaces. 

In the HDS framework, the LLM pre-training process itself constitutes the environment. To ensure the actor-critic agent can comprehensively perceive the training dynamics, the state is defined as a vector that captures the model's current performance, learning velocity, and internal stability. The agent's action corresponds to the sampling probabilities for data domains in the subsequent training batch. During inference, given the current state, HDS leverages the actor network within SAC to decide on this sampling probability distribution. To effectively encode the importance of each data domain from the state vector while maintaining computational efficiency, we designed compact, Transformer-like networks for both the actor and the critic.

The cornerstone of HDS is a holistic reward function engineered from three distinct perspectives. During training, we designed a multi-objective, holistic reward function for HDS that integrates three critical perspectives. The first component is a data-driven reward for data quality. The second is a gradient-driven reward for inter-domain influence, motivated by prior work like AC-ODM ~\cite{ma2025actor}. The third component is a model-centric reward based on the weight norms of specific LLM layers, designed to incentivize actions that promote rapid convergence and stability.

Our empirical results are compelling: as illustrated in Figure 1a, when pre-training Pythia-1B on The Pile, HDS reaches the final validation perplexity of the previous state-of-the-art method with 22\% fewer iterations and outperforms the static The Pile Weights (TPW) baseline by 57\% fewer iterations (Figure 1b). Furthermore, HDS achieves a 7.2\% improvement over the previous best method on the MMLU 0-shot task.

Our main contributions are as follows:
\begin{itemize}
    \item We propose the Holistic Data Scheduler (HDS), a novel online data mixing framework that employs the Soft Actor-Critic (SAC) algorithm. HDS introduces a multi-objective reward function that holistically integrates signals from data quality, training dynamics, and model stability, guided by efficiently designed actor-critic networks. This approach enables the discovery of more effective and stable data mixing policies.

    \item We establish HDS as the new state-of-the-art for non-proxy ODM techniques, demonstrating significant acceleration by matching the prior best perplexity with 22\% fewer iterations, alongside superior final performance with a 7.0\% perplexity reduction on the Pile test set and accuracy gains of 7.2\% (0-shot) and 4.0\% (5-shot) on MMLU, with consistent performance improvements on other downstream tasks.

    \item We conduct extensive ablation studies and provide a fine-tuned set of hyperparameters and design choices for HDS. This serves as a practical guide for researchers to effectively apply HDS to pre-train LLMs at even larger scales, facilitating broader adoption and future research.
\end{itemize}

\section{Methodology: The Holistic Data Scheduler}
\label{sec:method}

\begin{figure}[t]
  \centering
  \includegraphics[width=\linewidth]{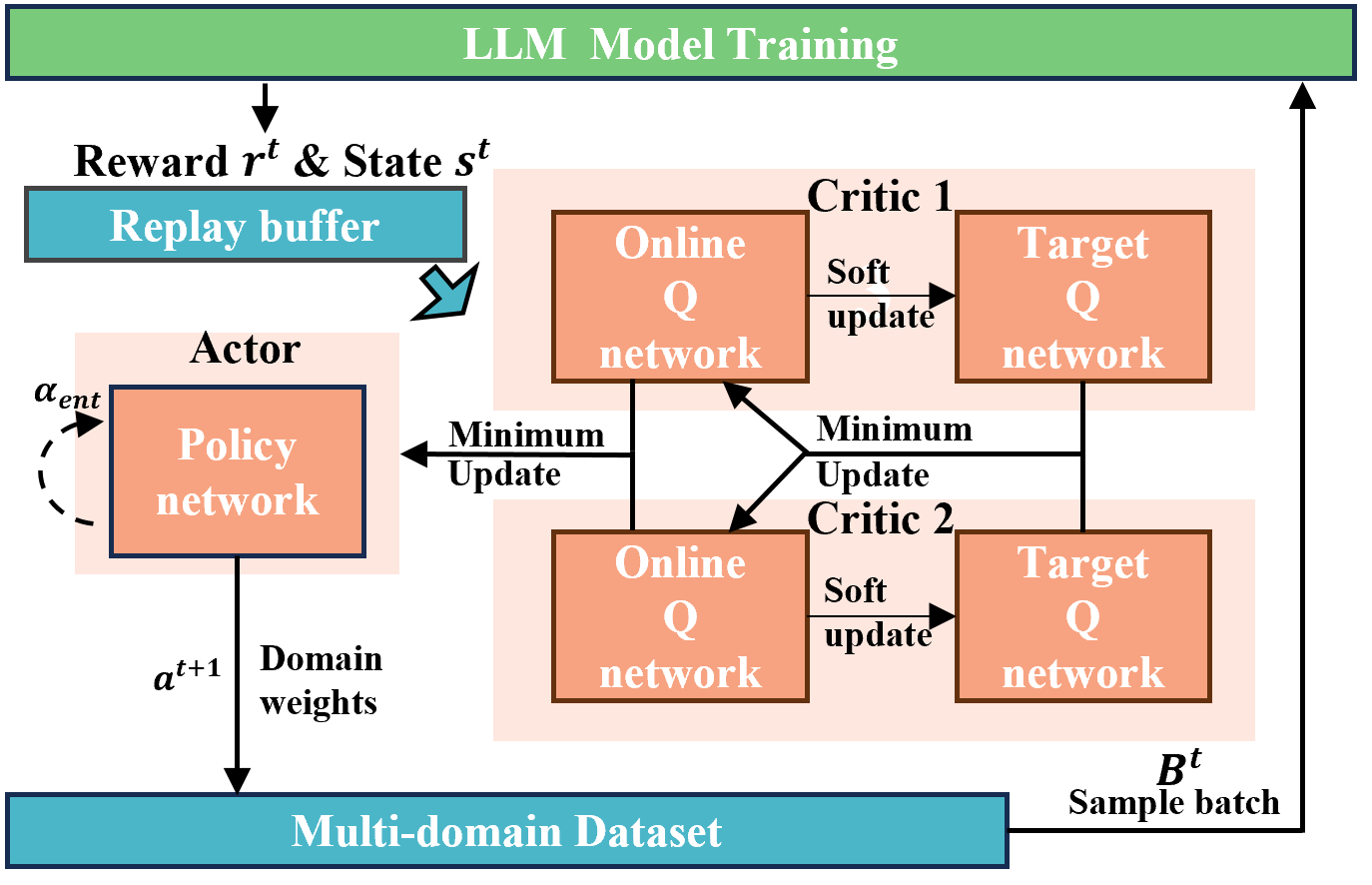}
  \caption{An overview of the Holistic Data Scheduler (HDS) framework. At each training step $t$, the SAC agent observes the state $s^t$ from the LLM environment. The actor network outputs a stochastic action $a^t$, which defines the domain weights for sampling the next data batch $B^t$. The LLM is then trained on $B^t$, leading to a new state $s_{t+1}$ and a holistic reward $r_t$. The transition tuple is stored in the experience replay buffer. The actor and critic networks, both with Transformer-based architectures, are updated by sampling mini-batches from the buffer. This cycle enables HDS to dynamically optimize the data mixing strategy throughout the pre-training process.}
  \Description{An overview of the Holistic Data Scheduler (HDS) framework.}
  \label{fig:arch}
\end{figure}

In this section, we introduce the \textbf{H}olistic \textbf{D}ata \textbf{S}cheduler (HDS), a novel online data mixing framework for LLM pre-training. At its core, HDS formulates the dynamic data scheduling problem as a continuous control task optimized via multi-objective reinforcement learning. We employ the Soft Actor-Critic (SAC) algorithm, renowned for its stability and sample efficiency, to navigate the high-dimensional policy space of data mixing strategies. The overall architecture, illustrating the interaction between the SAC agent and the LLM environment, is depicted in Figure ~\ref{fig:arch}.

\subsection{Problem Formulation}
\label{sec:problem_formulation}

Let $\mathcal{D} = \{D_1, D_2, \dots, D_K\}$ denote a large-scale pre-training corpus composed of $K$ distinct data domains. Our objective is to dynamically determine the optimal mixing weights for these domains at each training step. We define the domain weight vector at training step $t$ as $\bm{a}^t = [a_1^t, a_2^t, \dots, a_K^t] \in \Delta^K$, where $\Delta^K$ is the probability simplex in $\mathbb{R}^K$ (i.e., $\sum_{i=1}^K a_i^t = 1$ and $a_i^t \ge 0$).

The data mixture used to train the LLM is constructed based on these weights. Specifically, a training batch $B^t$ is formed by first sampling a domain $D_i$ with probability $a_i^t$, and then uniformly sampling a mini-batch of instances from that domain, denoted as $B_i^t \sim \text{UNIFORM}(D_i)$. This procedure induces an instance-wise sampling distribution over the entire corpus, given by $P_{\bm{a}^t} = \sum_{i=1}^K a_i^t \cdot \text{UNIFORM}(D_i)$.

In contrast to offline methods that fix $\bm{a}$ before training, our online approach aims to find an optimal sequence of policies $\{\pi_t\}_{t=0}^T$ that generate a sequence of domain weights $\{\bm{a}^t\}_{t=0}^T$. This sequence dynamically adapts to the LLM's learning state to maximize a cumulative, multi-objective reward signal, thereby accelerating convergence and enhancing final model performance.

\subsection{Online Data Mixing as a Multi-Objective RL Problem}
\label{sec:adapting_sac}

We formalize the dynamic data scheduling task as a Markov Decision Process (MDP), defined by the tuple $(\mathcal{S}, \mathcal{A}, \mathcal{P}, \mathcal{R}, \gamma)$, where $\mathcal{S}$ is the state space, $\mathcal{A}$ is the action space, $\mathcal{P}$ is the state transition function implicitly defined by the LLM training dynamics, $\mathcal{R}$ is our novel reward function, and $\gamma$ is the discount factor.

\paragraph{State Space ($\mathcal{S}$)} The state $\bm{s}^t \in \mathcal{S}$ at step $t$ is designed to provide the RL agent with a comprehensive snapshot of the LLM's training dynamics. It is a high-dimensional vector comprising six key components:
\begin{enumerate}
    \item \textbf{Sample Count ($\bm{n}^t$):} A vector of the number of samples drawn from each domain up to step $t$.
    \item \textbf{Training Step ($t$):} The current global training iteration.
    \item \textbf{Domain Losses ($\bm{l}^t$):} A vector of the validation loss for each domain.
    \item \textbf{Loss Deltas ($\Delta\bm{l}^t$):} The change in domain losses from the previous step, $\bm{l}^t - \bm{l}^{t-1}$.
    \item \textbf{Weight Norm ($||\bm{\omega}^t||_2$):} The $L_2$-norm of the weights of pre-selected layers in the LLM.
    \item \textbf{Weight Norm Delta ($||\Delta\bm{\omega}^t||_2$):} The change in the $L_2$-norm of weights from the previous step.
\end{enumerate}
This comprehensive state representation, $\bm{s}^t = (\bm{n}^t, t, \bm{l}^t, \Delta\bm{l}^t, ||\bm{\omega}^t||_2,\\ ||\Delta\bm{\omega}^t||_2)$, enables the agent to make informed decisions based on both historical trends and the instantaneous state of the LLM.

\paragraph{Action Space ($\mathcal{A}$)} The action $\bm{a}^t \in \mathcal{A}$ corresponds to the domain weight vector $\bm{a}^{t+1}$ to be used for sampling the next training batch. The action space is continuous and constrained to the probability simplex $\Delta^K$.

\paragraph{Policy ($\pi$)} HDS employs a stochastic policy $\pi_{\theta_A}(\bm{a}^t|\bm{s}^t)$, parameterized by the actor network with weights $\theta_A$. This policy outputs a probability distribution over the continuous action space, from which a specific domain weight vector $\bm{a}^{t+1}$ is sampled.

\subsection{Holistic Reward Function Design}
\label{sec:reward_design}

A pivotal innovation of HDS is its multi-objective reward function, which synthesizes three distinct perspectives to create a holistic training signal. The final reward for domain $i$ at step $t$, $r_i^t$, is a weighted combination of these components.

\paragraph{1. Inter-Domain Influence Reward ($r_{\text{align}}$)}
Inspired by DoGE~\cite{fan2024doge} and AC-ODM~\cite{ma2025actor}, this reward component quantifies the positive influence of training on one domain's data on all other domains. Let $\bm{g}_i^t = \nabla_{\theta_M} \mathcal{L}(\theta_M^t, B_i^t)$ be the gradient of the loss with respect to the LLM parameters $\theta_M$ for a batch $B_i^t$ from domain $i$. The alignment reward for domain $i$ is defined as the inner product of its gradient with the aggregated gradients of all other domains:
\begin{equation}
r_{\text{align}, i}^t = \left\langle \bm{g}_i^t, \sum_{j=1, j \neq i}^K \bm{g}_j^t \right\rangle
\end{equation}
A high $r_{\text{align}, i}^t$ indicates that updates from domain $i$ align well with the learning objectives of other domains, thus promoting the selection of data with high cross-domain utility and accelerating overall convergence.

\paragraph{2. Scheduled Lexical Diversity Reward ($r_{\text{diversity}}$)}
This component is designed to dynamically guide the model's exposure to textual complexity in alignment with its learning progression. We use the Measure of Textual Lexical Diversity (MTLD)~\cite{mccarthy2010mtld} as a proxy for syntactic complexity. The reward is defined to encourage the model to learn from simpler texts (lower MTLD) in early training stages and gradually shift to more complex texts (higher MTLD) as training progresses.

First, we compute the MTLD score for the batch from each domain, $\text{MTLD}(B_i^t)$. This score is normalized to $[0, 1]$:
\begin{equation}
\text{MTLD}_{\text{norm}}(B_i^t) = \frac{\text{MTLD}(B_i^t) - \text{MTLD}_{\min}}{\text{MTLD}_{\max} - \text{MTLD}_{\min}}
\end{equation}
where $\text{MTLD}_{\max}$ is the sequence length and $\text{MTLD}_{\min}=2$. We also normalize the current training step $t' = t / T_{\text{total}}$. The scheduled diversity reward is then:
\begin{equation}
r_{\text{diversity}, i}^t = \frac{t'}{\text{MTLD}_{\text{norm}}(B_i^t) + \epsilon}
\end{equation}
where $\epsilon$ is a small constant to prevent division by zero. This formulation creates a dynamic target: early on (small $t'$), the agent is rewarded for picking low-MTLD domains; later (large $t'$), the reward is higher for high-MTLD domains, creating a natural curriculum.

\begin{figure}[t]
  \centering
  \includegraphics[width=\linewidth]{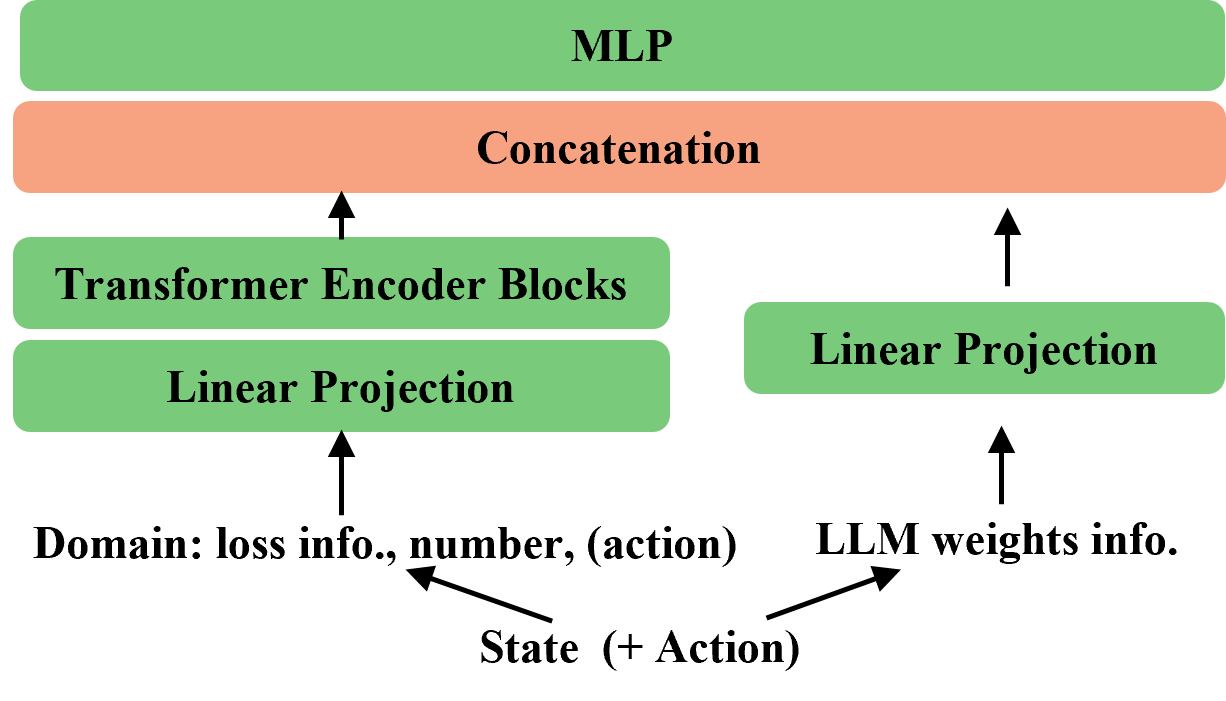}
  \caption{The network architecture for the actor and critic. The action is excluded from the input for the actor network.}
  \Description{The network architecture for the actor and critic.}
  \label{fig:ac}
\end{figure}

\paragraph{3. Model Stability Reward ($r_{\text{stability}}$)}
This reward promotes smooth and stable convergence by penalizing drastic fluctuations in the model's parameters. It is defined as the inverse of the magnitude of change in the $L_2$-norm of selected layer weights $\bm{\omega}^t$ (the same layers used for state representation) and we use $ \hat{r}_{\text{stability}}$ to prevent it from getting too large late in training:
\begin{equation}
r_{\text{stability}}^t = max(\frac{1}{\left| ||\bm{\omega}^t||_2 - ||\bm{\omega}^{t-1}||_2 \right| + \epsilon}, \hat{r}_{\text{stability}})
\end{equation}
This scalar reward is applied uniformly across all domains. It incentivizes the agent to choose data mixtures that lead to steady learning, avoiding overly aggressive updates that might destabilize the training process.

\paragraph{Final Reward} The total reward vector $\bm{r}^t$ is a weighted sum of these three components:
\begin{equation}
\bm{r}^t = w_{\text{align}} \bm{r}_{\text{align}}^t + w_{\text{diversity}} \bm{r}_{\text{diversity}}^t + w_{\text{stability}} r_{\text{stability}}^t \cdot \mathbf{1}
\end{equation}
where $w_{(\cdot)}$ are hyperparameters balancing the contribution of each objective, and $\mathbf{1}$ is a vector of ones which the dimension is the numbers of domains.

\begin{algorithm}[t]
\caption{The Holistic Data Scheduler (HDS) Algorithm}
\label{alg:hds}
\small
\SetKwInOut{Input}{Input}
\SetKwInOut{Initialize}{Initialize}

\Input{Corpus $\mathcal{D}=\{D_1, ..., D_K\}$, hyperparameters}
\Initialize{Actor $\pi_{\theta_A}$, critics $Q_{\theta_{C1,C2}}$, target critics $Q_{\theta'_{C1,C2}}$ with $\theta' \leftarrow \theta$, replay buffer $\mathcal{B}$, number of training 
 $N_{ac}$, entropy temperature $\alpha_{\text{ent}}$, LLM $\theta_M$.}

\For{$t=1, \dots, T$}{
    Observe state $\bm{s}^t$ from the LLM environment\;
    Sample action $\bm{a}^t$\;
    Sample data batch $B^t$ according to $P_{\bm{a}^{t+1}}$\;
    Update LLM parameters: \\
    \quad $\theta_M^{t+1} \leftarrow \theta_M^t - \eta_M \sum_{i=1}^K a_i^{t+1} \nabla \mathcal{L}(\theta_M^t, B_i^t)$\;
    Compute reward vector $\bm{r}^t$ using Eq. (1)-(5)\;
    Observe next state $\bm{s}^{t+1}$\;
    Store transition $(\bm{s}^t, \bm{a}^t, \bm{r}^t, \bm{s}^{t+1})$ in $\mathcal{B}$\;
    
    \For{$n_{ac}=1, \dots, N_{ac}$}{
        Sample a minibatch $\{(\bm{s}^j, \bm{a}^j, \bm{r}^j, \bm{s}^{j+1})\}$ from $\mathcal{B}$\;
        Compute target $y^j$ for each transition in the minibatch\;
        Update critics by minimizing $\mathcal{L}(\theta_{C1})$ and $\mathcal{L}(\theta_{C2})$\;
        Update actor by minimizing $\mathcal{L}(\theta_A)$\;
        Update entropy temperature by minimizing $\mathcal{L}(\alpha_{\text{ent}})$\;
        Update target networks using Polyak averaging\;
    }
}
\end{algorithm}

\subsection{HDS Agent and Model Update}
\label{sec:model_update}
Each iteration involves updating the LLM parameters $\theta_M$ and the HDS agent's parameters (actor $\theta_A$, critics $\theta_{C1}, \theta_{C2}$, and entropy temperature $\alpha_{\text{ent}}$).

\paragraph{LLM Update}
Given the domain weights $\bm{a}^t$ sampled from the policy, the LLM parameters are updated via stochastic gradient descent:
\begin{equation}
\theta_M^{t+1} \leftarrow \theta_M^t - \eta_M \sum_{i=1}^K a_i^t \nabla_{\theta_M} \mathcal{L}(\theta_M^t, B_i^t)
\end{equation}
where $\eta_M$ is the learning rate for the LLM.

\paragraph{SAC Agent Update}
The SAC agent learns from transitions $(\bm{s}^t, \bm{a}^t,\\ \bm{r}^t, \bm{s}^{t+1})$ stored in a replay buffer $\mathcal{B}$.
The update process leverages twin Q-networks (critics) to mitigate overestimation bias.

\textbf{Critic Update.} The two critic networks, $Q_{\theta_{C1}}$ and $Q_{\theta_{C2}}$, are updated by minimizing the soft Bellman residual. The target value $y$ for a transition $(\bm{s}, \bm{a}, \bm{r}, \bm{s}')$ is:
\begin{equation}
y = \bm{r} + \gamma \left( \min_{k=1,2} Q_{\theta'_{Ck}}(\bm{s}', \bm{a}') - \alpha_{\text{ent}} \log \pi_{\theta_A}(\bm{a}'|\bm{s}') \right)
\end{equation}
where $\bm{a}' \sim \pi_{\theta_A}(\cdot|\bm{s}')$, and $\theta'_{Ck}$ are the weights of the target critic networks. The loss for each critic is the mean squared error:
\begin{equation}
\mathcal{L}(\theta_{Ck}) = \mathbb{E}_{(\bm{s},\bm{a},\bm{r},\bm{s}') \sim \mathcal{B}} \left[ \left( Q_{\theta_{Ck}}(\bm{s},\bm{a}) - y \right)^2 \right], \quad k=1,2
\end{equation}

\begin{figure*}[t]
    \centering
    \includegraphics[width=\textwidth]{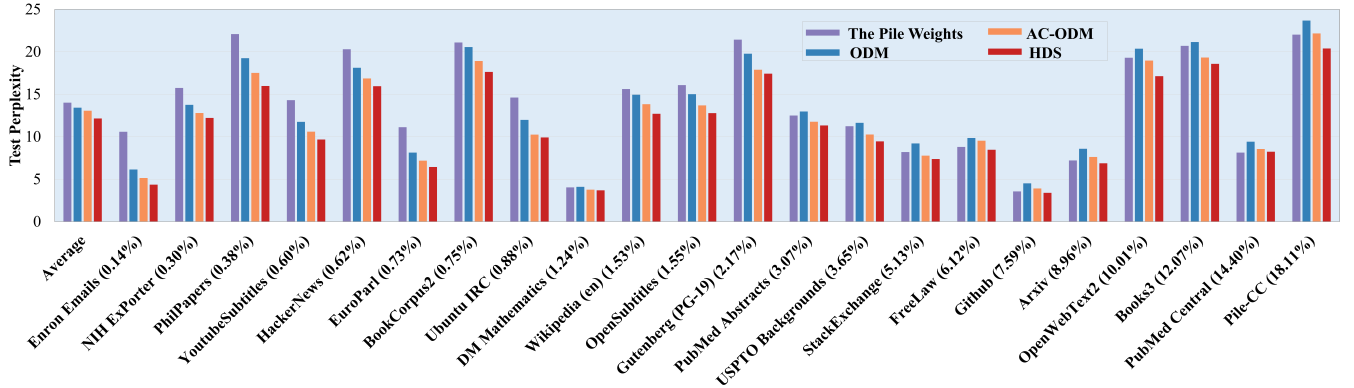}
    \caption{Test perplexity on the average and across the 22 individual domains of The Pile. The horizontal axis indicates the domain name and its corresponding token proportion in the training set. HDS consistently achieves lower perplexity across nearly all domains, demonstrating its robustness and ability to effectively balance learning dynamics.}
    \Description{Test perplexity on the average and across the 22 individual domains of The Pile.}
    \label{fig:domain_ppl}
\end{figure*}

\textbf{Actor and Entropy Update.} The actor network is updated by maximizing the expected return and the policy entropy. The policy loss is:
\begin{equation}
\mathcal{L}(\theta_A) = \mathbb{E}_{\bm{s} \sim \mathcal{B}, \bm{a} \sim \pi_{\theta_A}} \left[ \alpha_{\text{ent}} \log \pi_{\theta_A}(\bm{a}|\bm{s}) - \min_{k=1,2} Q_{\theta_{Ck}}(\bm{s},\bm{a}) \right]
\end{equation}
The entropy temperature $\alpha_{\text{ent}}$ is also learned by minimizing its own loss function to dynamically balance the reward and policy entropy:
\begin{equation}
\mathcal{L}(\alpha_{\text{ent}}) = \mathbb{E}_{\bm{a} \sim \pi_{\theta_A}} [-\alpha_{\text{ent}} \log \pi_{\theta_A}(\bm{a}|\bm{s}) - \alpha_{\text{ent}} \mathcal{H}]
\end{equation}
where $\mathcal{H}$ is a target entropy hyperparameter.

\textbf{Network Architecture.} As shown in Figure~\ref{fig:ac}ur actor and critic networks employ a compact, Transformer-based architecture. This design leverages self-attention mechanisms to effectively model the complex inter-dependencies within the high-dimensional state vector $\bm{s}^t$, enabling a more nuanced mapping from state to action and value.

\textbf{Target Network Update.} The target critic networks are updated via Polyak averaging:
\begin{equation}
\theta'_{Ck} \leftarrow \tau \theta_{Ck} + (1-\tau) \theta'_{Ck}
\end{equation}
where $\tau \ll 1$ is the target smoothing coefficient.

To summarize, the training procedure of HDS is described in Algorithm~\ref{alg:hds}.

\section{Experiments and Analysis}
In this section, we conduct a series of experiments to rigorously evaluate the performance of our proposed Holistic Data Scheduler (HDS). We first detail the experimental setup, then present the main results, and finally provide in-depth analyses on performance improvements and computational efficiency.

\subsection{Experimental Setup}
Our experimental framework is designed to ensure reproducibility and fair comparison. All experiments were conducted on a consistent hardware platform equipped with an Intel Xeon Platinum 8468 CPU and eight NVIDIA H800 80GB GPUs.

\paragraph{Dataset and LLM Training}
We use The Pile~\cite{gao2020pile}, a large-scale, diverse, and open-source dataset for language model pre-training. It consists of 825GB of text from 22 different domains. For our experiments, we train a 1-billion-parameter decoder-only Transformer model based on the Pythia suite~\cite{biderman2023pythia}, leveraging a modified version of the GPT-NeoX library~\cite{black2022gpt}. The model is trained for 50 billion tokens, which corresponds to 41,667 training steps. We set the sequence length to 1024 and utilize sequence packing to improve efficiency. The global batch size is 1152, achieved through a micro-batch size of 8 per GPU and gradient accumulation over 18 steps. During a warm-up phase of the first 833 steps, we apply the original domain weights from The Pile with a small Gaussian noise $N(0, 0.02)$ to stabilize initial training, before activating the HDS policy.

\paragraph{SAC Setting}
The core of HDS is the Soft Actor-Critic (SAC) agent, which dynamically adjusts domain weights. For the model-driven reward component and state representation, we select the weights from the Transformer layers at even-numbered indices, as well as the first layer, to balance representational richness and computational cost. To ensure the three reward components contribute on a comparable scale, we set their respective weights as $w_{\text{align}}=1$, $w_{\text{diversity}}=10$, and $w_{\text{stability}}=10$. Furthermore, the stability reward is capped at an upper limit of 5 to prevent extreme values from destabilizing the agent's learning process.

The actor and critic networks are crucial for the agent's performance. For our 1B parameter LLM, we designed lightweight yet powerful network architectures for both. They consist of a linear projection layer with a dimension of 256, followed by 8 Transformer encoder blocks with a hidden dimension of 512. The final output is produced by a 4-layer MLP. This architecture results in a total of approximately 5 million parameters for each network, ensuring minimal computational overhead while effectively processing the complex state information. And $N_{ac}$ is set to 2.

\paragraph{Baselines}
We compare HDS against several strong baselines to demonstrate its effectiveness:
\begin{itemize}
    \item \textbf{The Pile Weights (TPW)}: The static, heuristic-based domain weights originally provided with The Pile dataset.
    \item \textbf{ODM}~\cite{albalak2023efficient}: An online data mixing method that models the problem as a multi-armed bandit, using domain-specific loss as the reward signal.
    \item \textbf{AC-ODM}~\cite{ma2025actor}: An advanced online method using an actor-critic framework that considers inter-domain interactions through a gradient alignment reward.
\end{itemize}

\paragraph{Evaluation}
Our evaluation is twofold. First, we measure the model's language modeling capability using the unweighted average perplexity on the validation sets of all 22 domains in The Pile. Second, to assess downstream task performance, we evaluate the models on the MMLU benchmark~\cite{hendrycks2020measuring}, which covers 57 diverse subjects. We report both 0-shot and 5-shot accuracy.

\subsection{Main Results}

Our primary findings, illustrated in Figure~\ref{fig:f1}, demonstrate the significant advantages of HDS in both training efficiency and final model performance.

As shown in Figure~\ref{fig:f1a}, HDS exhibits a substantially steeper decline in validation perplexity compared to all baselines. It reaches the final perplexity level of the strongest baseline, AC-ODM, with approximately 44\% fewer training steps. When compared to the static TPW baseline, HDS achieves the same perplexity with a remarkable 57\% reduction in training steps. By the end of the training at 41,667 steps, HDS achieves a final validation perplexity that is 13.6\% lower than TPW, 8.9\% lower than ODM, and 5.3\% lower than AC-ODM. This highlights the superior efficiency and effectiveness of our multi-objective optimization approach, which enables the model to learn more effectively from the data mixture at every stage of training. Beyond these perplexity improvements, we observe a strong correlation with downstream capability gains.

Complementing the perplexity results, Figure~\ref{fig:f1b} presents the evolution of 0-shot accuracy on the MMLU benchmark throughout the pre-training process. While the AC-ODM baseline exhibits competitive performance in the earliest stages, HDS demonstrates a more rapid and sustained trajectory of improvement. Crucially, HDS surpasses all baselines by the 20k step mark and maintains a distinct lead until the end of training. This consistent superiority in test accuracy indicates that the lower perplexity achieved by HDS translates directly into robust, generalizable representations, rather than merely overfitting to the validation set language modeling objective.

\begin{table}[htbp]
  \centering
  \caption{Evaluation of downstream task performance on the MMLU benchmark. Accuracy (Acc) is reported.}
  \label{tab:mmlu_results}
  \begin{tabular}{lcc}
    \toprule
    \textbf{Algorithm} & \textbf{0-shot} & \textbf{5-shot} \\
    \midrule
    TPW                & 0.20664                    & 0.27469                    \\
    ODM                & 0.23514                    & 0.28416                    \\
    AC-ODM             & 0.25146                    & 0.29868                    \\
    \textbf{HDS}       & \textbf{0.26915}           & \textbf{0.31064}           \\
    \bottomrule
  \end{tabular}
\end{table}

The downstream performance on the MMLU benchmark, detailed in Table~\ref{tab:mmlu_results}, further corroborates these findings. HDS not only trains faster but also produces a more capable model. In the 0-shot setting, HDS achieves an accuracy of 0.26915, outperforming all baselines. Similarly, in the 5-shot setting, HDS attains an accuracy of 0.31064. This consistent improvement on complex downstream tasks underscores that the holistic reward function in HDS guides the pre-training process towards developing more generalizable and robust representations.

\begin{table}[htbp]
\caption{Zero shot accuracy on downstream tasks using The Pile pretrained Pythia-1B models.}
\label{tab:downstream-six-transposed}
\centering
\begin{tabular}{lcccc}
\toprule
Method & \textbf{COPA} & \textbf{LogiQA} & \textbf{PIQA} & \textbf{WinoGrande} \\
\midrule
TPW & 0.54800 & 0.23810 & 0.60330 & 0.50930 \\
ODM & 0.68000 & \textbf{0.30720} & 0.68330 & 0.59650 \\
AC-ODM & 0.69800 & 0.30110 & 0.69670 & 0.58300 \\
\textbf{HDS} & \textbf{0.71000} & 0.30550 & \textbf{0.69840} & \textbf{0.65120} \\
\bottomrule
\end{tabular}
\end{table}

The zero-shot performance on specific downstream tasks, presented in Table~\ref{tab:downstream-six-transposed}, provides additional evidence of the effectiveness of our approach. HDS demonstrates superior generalization capabilities, outperforming baselines on three out of the four evaluated datasets. Specifically, HDS achieves the highest accuracy on COPA (0.71000), PIQA (0.69840), and WinoGrande (0.65120), significantly surpassing TPW and AC-ODM. While ODM yields a marginal advantage on LogiQA, HDS remains highly competitive with a comparable score of 0.30550. This robust performance across diverse reasoning and commonsense tasks reinforces the conclusion that the HDS data mixing strategy effectively facilitates the acquisition of more transferable knowledge during pre-training.

\subsection{Reduction of Perplexity to Baselines}

The holistic reward function of HDS is engineered to create a synergistic learning curriculum by dynamically balancing data quality, inter-domain influence, and model stability. This allows HDS to exploit shared knowledge across domains, thereby accelerating the learning process. Figure~\ref{fig:domain_ppl} presents the test perplexity for each method on held-out data from all 22 domains, providing a granular view of their performance.

A critical component driving this robust performance is the scheduled lexical diversity reward ($r_{\text{diversity}}$), which introduces an explicit curriculum learning dynamic absent in prior methods. This reward component incentivizes the model to first learn from domains with lower lexical diversity (i.e., simpler, more repetitive language structures) during the initial training phases. As the model's capabilities grow, the reward schedule shifts to prioritize domains with higher lexical diversity (i.e., more complex and varied vocabulary and syntax). This "simple-to-complex" progression prevents the model from being overwhelmed by difficult text early on and ensures it builds a solid linguistic foundation. For instance, domains like Ubuntu IRC or OpenSubtitles, characterized by conversational and often less complex text ~\cite{gao2020pile}, are reduced little. Subsequently, the model is better prepared to tackle the rich, formal, and diverse content found in domains like FreeLaw or BookCorpus2~\cite{gao2020pile}, leading to more efficient and stable learning in final stages.

This curriculum-aware scheduling explains why HDS demonstrates superior or highly competitive performance across the vast majority of domains, in contrast to methods that might excel in some areas but falter in others. For instance, in domains with small to medium data proportions (e.g., NIH ExPorter, Ubuntu IRC, DM Mathematics), where efficient knowledge extraction is critical, HDS shows a clear advantage. Its reward mechanism not only identifies high-utility data but also sequences it appropriately based on its complexity.

Furthermore, in large-data domains (e.g., Pile-CC, PubMed Central, Books3), which form the backbone of the pre-training corpus, HDS also consistently outperforms other methods. This indicates that the scheduler does not merely focus on smaller, more challenging domains but maintains a global, curriculum-informed perspective, optimizing the data mixture to ensure robust learning across the entire data distribution. The ability of HDS to dynamically leverage the strengths of all domains by prioritizing samples that maximize cross-domain knowledge transfer and model stability, all while respecting the model's current learning stage, is a key factor behind its state-of-the-art performance. This validates the robustness of our multi-objective approach in complex, multi-domain pre-training scenarios.

\begin{table*}[!t]
  \centering
  \caption{Comparison of model size and computational cost during pre-training. The columns AC, LLM, and AC+LLM show the parameter counts for the agent, the language model, and the combined system. Time per step, total steps required to match the final perplexity of the Pile weights (TPW), and the resulting speedup ratio are also reported.}
  \label{tab:computation_cost}
  \begin{tabular}{lcccccc}
    \toprule
    \textbf{Algorithm} & \textbf{AC (Params)} & \textbf{LLM (Params)} & \textbf{AC+LLM (Params)} & \textbf{Time per Step (s)} & \textbf{Steps} & \textbf{Speedup Ratio} \\
    \midrule
    TPW   & 0                    & 1B                    & 1B                       & 2.47                       & 41667          & 1.00x                  \\
    ODM                & 0                    & 1B                    & 1B                       & 2.47                       & 29162          & 1.37x                  \\
    AC-ODM             & 17.32M               & 1B                    & 1.02B                    & 2.48                       & 21330          & 1.87x                  \\
    \textbf{HDS}       & \textbf{26.5M}       & \textbf{1B}           & \textbf{1.03B}           & \textbf{2.49}              & \textbf{17917} & \textbf{2.21x}         \\
    \bottomrule
  \end{tabular}
\end{table*}

\subsection{Computational Cost Reduction Relative to Others}
To provide a comprehensive assessment of efficiency, we analyze the computational cost required for each method to train a 1B LLM to a comparable level of performance. We use the final perplexity achieved by the TPW baseline as the convergence target and measure the steps and time taken by each method to reach it.

The results, summarized in Table~\ref{tab:computation_cost}, highlight the substantial computational savings offered by HDS. While introducing an intelligent scheduling agent (the SAC actor and critic) adds a small number of parameters (26.5M, a $\sim$2.6\% increase), the operational overhead is minimal. The time per step increases only marginally from 2.47s to 2.49s—an increase of less than 1\%—indicating that the integration of HDS does not impose a bottleneck on the training throughput.

HDS reaches the target perplexity in just 17,917 steps. This is a significant reduction compared to the static TPW baseline, as well as the dynamic ODM and AC-ODM methods. This translates to a remarkable speedup ratio of 2.21x relative to the baseline training time. This efficiency gain stems directly from the superior sample quality provided by the HDS policy at each step, allowing the model to learn faster and more effectively.

Furthermore, we analyze the implications of these findings for scaling to larger model architectures. The computational overhead of the HDS agent is primarily driven by its forward pass, which is computationally inexpensive compared to the gradient-heavy backpropagation of the LLM. As indicated by our hyperparameter studies, the agent's size does not need to scale linearly with the backbone model and remains effective at a small fraction (e.g., $\approx$0.5\%) of the LLM's parameters. Consequently, as the target model scales to larger dimensions (e.g., 7B or beyond), the relative cost of the agent is expected to become asymptotically negligible. This suggests that the substantial speedup ratios observed here are likely to persist or even improve in larger-scale scenarios, validating HDS as a highly cost-effective solution for resource-intensive foundation model pre-training.

\begin{figure}[t]
  \centering
  \includegraphics[width=\columnwidth]{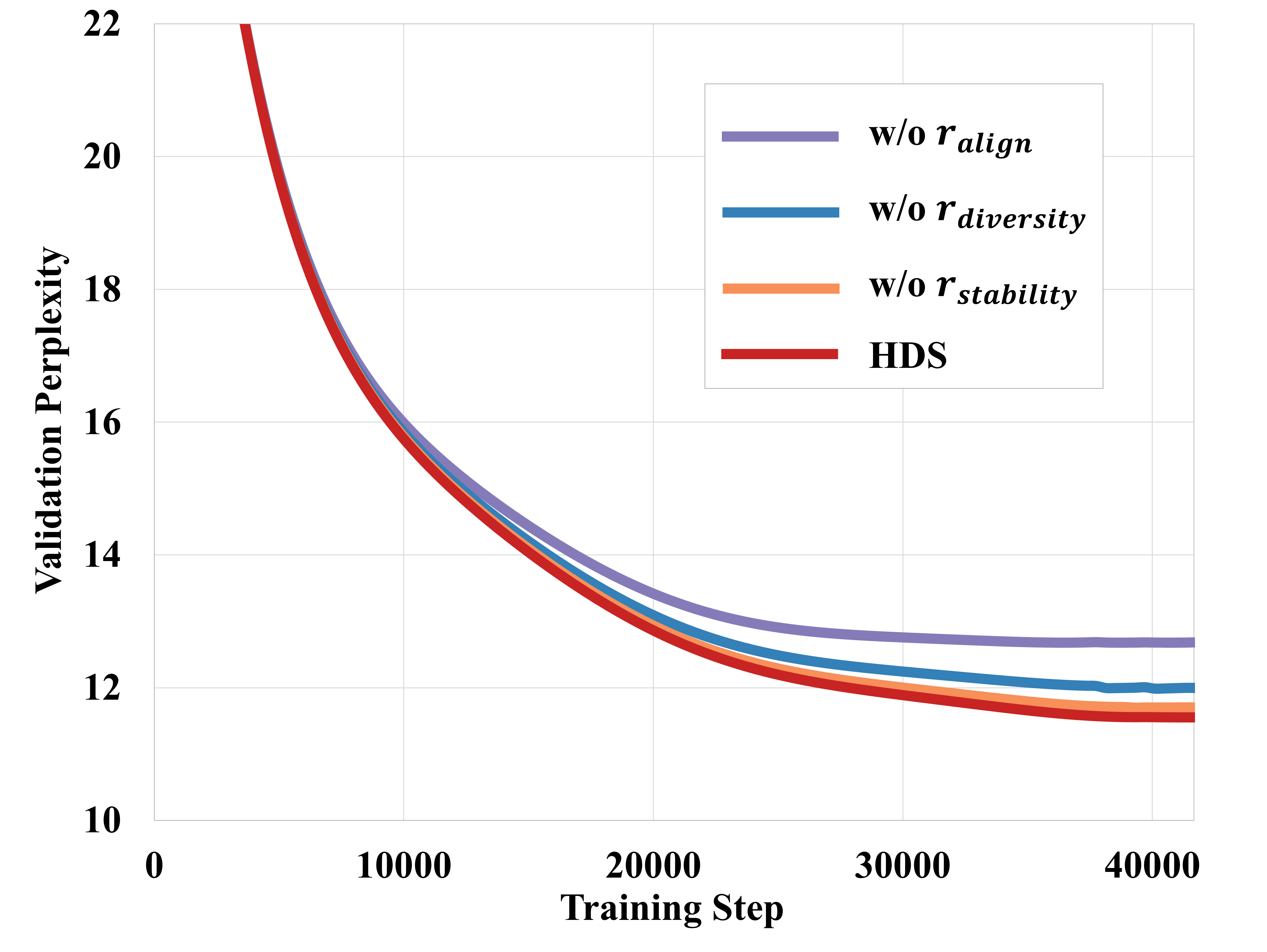}
  \caption{Ablation study of the reward components. The validation perplexity curves show the performance of the full HDS model against variants with each of the three reward signals removed (w/o denotes "without").}
  \label{fig:ablation}
  \Description{Ablation study of the reward components.}
\end{figure}

\subsection{Ablation Study on Reward Components}
To dissect the individual contributions of our multi-objective reward function, we conducted a comprehensive ablation study. We systematically removed each of the three reward components—inter-domain influence ($r_{\text{align}}$), scheduled lexical diversity ($r_{\text{diversity}}$), and model stability ($r_{\text{stability}}$)—and evaluated the impact on the pre-training performance. For a more granular analysis concerning the sensitivity of the specific weights assigned to these reward components, please refer to Appendix~\ref{sec:appendix_reward_weights}. The results, as illustrated in Figure~\ref{fig:ablation}, reveal the critical and synergistic roles played by each component.

The full HDS model, which integrates all three rewards, consistently achieves the lowest validation perplexity (red curve), confirming that the components work in concert to produce the best results. The most significant performance degradation occurs when the inter-domain influence reward is removed (w/o $r_{\text{align}}$). As shown by the purple curve in Figure~\ref{fig:ablation}, the final perplexity of this variant is substantially higher than all others, underscoring that $r_{\text{align}}$ is the cornerstone of our framework. The widening gap between this variant and the full model throughout training suggests that without prioritizing data that maximizes positive gradient alignment, the model struggles to capture the underlying synergies between domains, severely hampering acceleration and knowledge transfer.

\begin{figure}[t]
  \centering
  \includegraphics[width=\columnwidth]{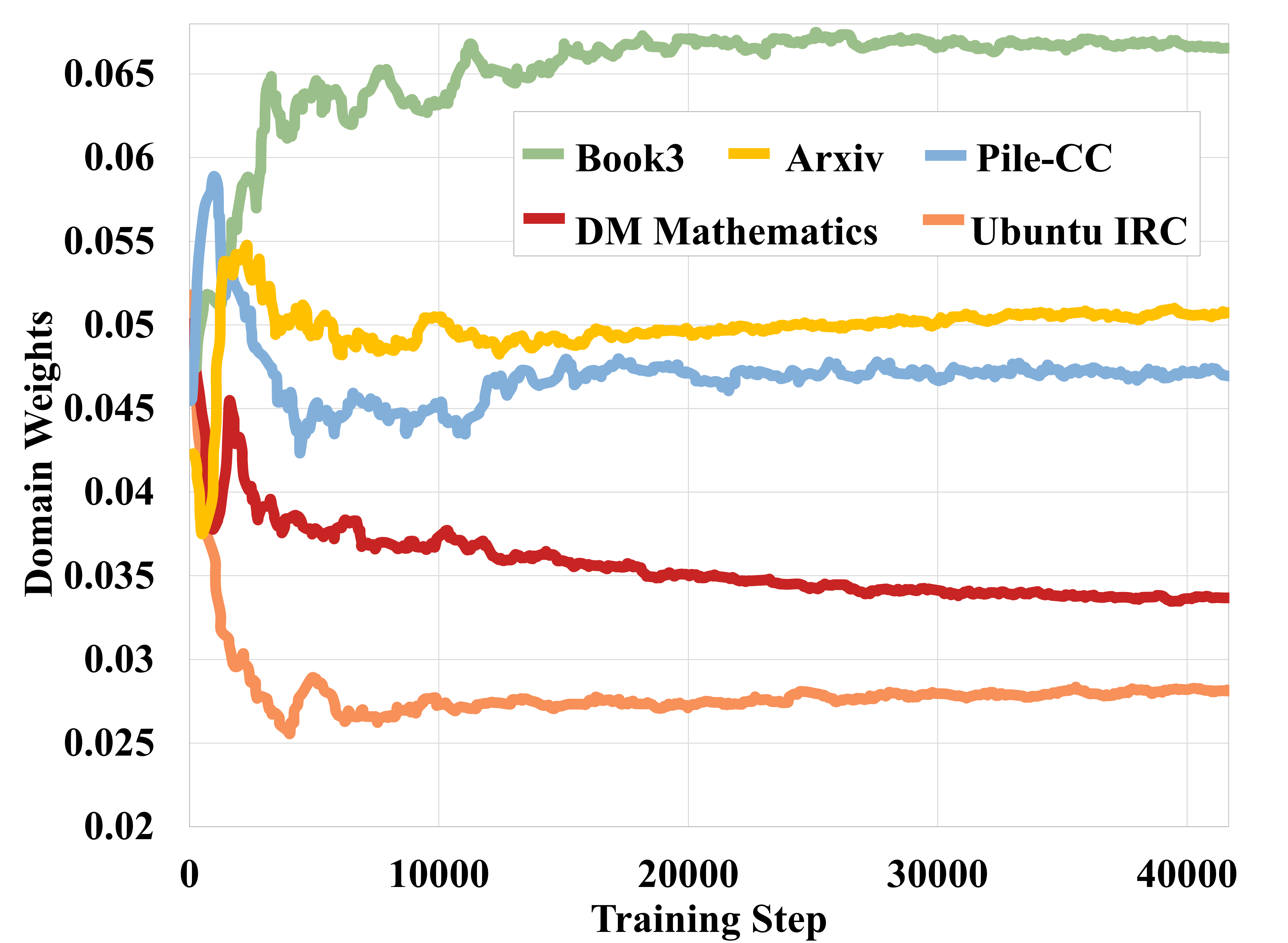}
  \caption{Evolution of domain weights for five selected domains with obvious changes during pre-training. The dynamic adjustments reflect the policy learned by HDS to optimize the data mixture.}
  \label{fig:domain_weights}
  \Description{Evolution of domain weights for five selected domains with obvious changes during pre-training.}
\end{figure}

Removing the scheduled lexical diversity reward (w/o $r_{\text{diversity}}$) leads to the second-largest decline in performance, represented by the blue curve. This result validates the efficacy of our explicit curriculum learning strategy. Unlike the stability component, the lack of $r_{\text{diversity}}$ results in a distinct performance gap that persists from the early stages to convergence. Without the "simple-to-complex" guidance provided by this reward, the model is less efficient at building its linguistic foundation, resulting in a slower convergence rate. This highlights the importance of not just selecting high-quality data, but also scheduling its delivery in alignment with the model's evolving capabilities.

Finally, the model trained without the stability reward (w/o $r_{\text{stability}}$) performs better than the other two ablated variants but still consistently falls short of the complete HDS model (orange curve vs. red curve). This indicates that while promoting smooth and steady updates via $r_{\text{stability}}$ is not as impactful as capturing inter-domain synergies or implementing a curriculum, it remains a valuable component. It acts as a regularizer, preventing erratic training dynamics that could otherwise slightly hinder the model's learning trajectory. In summary, this ablation study decisively demonstrates that all three reward components are integral to HDS's state-of-the-art performance, with each addressing a distinct and crucial aspect—alignment, curriculum, and stability—of the complex LLM pre-training process.

\begin{table}[t]
  \centering
  \caption{Final perplexity on The Pile test set for Pythia models of varying sizes, trained with HDS agents of different parameter counts (expressed as a percentage of the LLM's parameters).}
  \label{tab:optim}
  \begin{tabular}{l|ccccc}
    \toprule
    \textbf{Params} & \textbf{0.1\%} & \textbf{0.3\%} & \textbf{0.5\%} & \textbf{1.0\%} & \textbf{1.5\%}\\
    \midrule
    70M             & 22.86          & \textbf{21.24} & 21.28          & 21.24          & 21.58          \\
    160M            & 19.96          & 19.65          & 18.88          & 18.90          & \textbf{18.85} \\
    410M            & 14.95          & 14.56          & \textbf{14.33} & 14.34          & 14.33          \\
    1B              & 13.24          & 12.64          & \textbf{12.15} & 12.15          & \textbf{12.14} \\
    \bottomrule
  \end{tabular}
\end{table}

\subsection{Analysis of Dynamic Domain Weights}
To further understand the inner workings of our HDS framework, we visualize the evolution of the domain sampling weights throughout the pre-training process. Figure~\ref{fig:domain_weights} illustrates the dynamic weight adjustments for domains with obvious changes, offering insight into how HDS's policy adapts over time.

Initially, the weights for all domains exhibit sharp fluctuations as the SAC agent explores the policy space to learn the complex relationship between data mixture and model performance. Following this initial exploration phase, clear and meaningful trends emerge.

The weight for Ubuntu IRC, a domain characterized by conversational text and relatively low lexical diversity, quickly stabilizes at a low value. This aligns perfectly with our curriculum learning objective ($r_{\text{diversity}}$),: its simpler linguistic structures are most valuable during the early stages of training to build a foundational understanding, after which its marginal utility diminishes, and its sampling weight is appropriately reduced.

Conversely, the weight for Book3, a domain renowned for its high-quality, general-purpose, and lexically diverse text, shows a consistent upward trend, eventually becoming the most heavily weighted domain. This demonstrates the effectiveness of the $r_{\text{diversity}}$ component in action. As the model matures and training progresses (i.e., as $t'$ increases), the reward function increasingly prioritizes complex and diverse data to facilitate more advanced learning, correctly identifying Book3 as a prime source for this purpose.

The trajectories of the specialized, high-diversity domains like Arxiv and DM Mathematics are also revealing. Their weights initially spike, likely driven by high inter-domain influence ($r_{\text{align}}$) as the model tackles their unique vocabularies, before settling at a moderate, stable level. This suggests that while their specialized knowledge is consistently valuable, HDS learns to balance it against the need for broad, general-purpose data from domains like Book3 and Pile-CC.

\section{Hyperparameter optimization}

Given the immense computational and time costs associated with LLM pre-training, optimizing the hyperparameters of the HDS framework is crucial for its practical application. To this end, we conducted a series of targeted experiments to refine key design choices and provide empirical guidance for researchers aiming to apply HDS to larger-scale models. 

Our experiments on layer selection strategies for both the state's weight norm and the $r_{\text{align}}$ calculation (detailed in Appendix ~\ref{sec:appendix_c}) yielded clear insights. For the \textbf{state representation}, incorporating a larger number of the LLM's layers generally leads to a more comprehensive state capture and better performance, albeit at a higher computational cost. A layer-sampling approach (e.g., selecting all even- or odd-indexed layers) provides a highly effective trade-off, maintaining a global view of the training dynamics while improving efficiency. Conversely, for the \textbf{alignment reward ($r_{\text{align}}$)}, selecting layers from the latter half of the model tends to yield slightly better results. This is likely because deeper layers encode richer semantic information, which is more indicative of the cross-domain knowledge transfer we aim to optimize.

Furthermore, we rigorously investigated the optimal parameter count for the Actor and Critic networks relative to the LLM's size. The results, presented in Table~\ref{tab:optim}, reveal a consistent pattern across model scales. The analysis shows that performance generally improves as the agent's network size increases, but the gains follow a law of diminishing returns.  

For most model scales, an agent size of approximately \textbf{0.5\%} of the LLM's parameters strikes an excellent balance, delivering near-optimal performance without incurring unnecessary computational overhead.

Based on these findings, we offer the following practical guidelines for scaling HDS:
\begin{itemize}
    \item \textbf{State Representation:} Depending on hardware and time constraints, select a layer-sampling strategy (e.g., all even- or odd-indexed layers) to ensure the state's weight norm captures a broad, representative snapshot of the LLM's evolving weights.
    \item \textbf{Alignment Reward:} For the $r_{\text{align}}$ calculation, prioritize selecting a few layers from the latter half of the model, as these are more indicative of the model's higher-level semantic understanding and facilitate better gradient alignment.
    \item \textbf{Agent Network Size:} Choose an Actor and Critic network size that is approximately 0.3\% to 1.5\% of the target LLM's parameter count. A ratio of around \textbf{0.5\%} serves as a robust and efficient starting point that balances capability with training throughput.
\end{itemize}

\begin{table}[t]
  \centering
  \caption{Validation perplexity on The Pile during Pythia-12B pretraining.}
  \label{tab:val_perplexity}
  \begin{tabular}{lcccc}
    \toprule
        \textbf{Steps}
        & \textbf{5,208} & \textbf{10,416} & \textbf{15,624} & \textbf{20,832} \\
    \midrule
    ODM & 14.53 & 10.01 & 8.55 & 7.32 \\
    HDS & \textbf{13.46} & \textbf{8.19} & \textbf{6.04} & \textbf{4.89} \\
    \bottomrule
  \end{tabular}
\end{table}

To empirically validate these guidelines and assess the scalability of our approach, we conducted a large-scale pre-training experiment on the \textbf{Pythia-12B} model. For this evaluation, we maintained consistency with the training hyperparameters used in our main experiments, while the HDS agent's architecture was configured strictly according to the practical recommendations derived above (using $\approx$0.5\% parameter ratio). The model was trained for a total of 25 billion tokens on \textit{The Pile}.

The results, detailed in Table~\ref{tab:val_perplexity}, clearly demonstrate the effectiveness of HDS when scaled to larger architectures. HDS exhibits superior convergence properties, consistently achieving lower validation perplexity than the ODM baseline at every checkpoint. \textbf{Most notably, the performance gap widens significantly as training progresses.} At step 5,208, the perplexity difference is approximately 1.07; by the final checkpoint (20,832 steps), HDS attains a perplexity of \textbf{4.89}, significantly outperforming ODM's 7.32—a relative improvement of over 33\%. This substantial margin confirms that the holistic reward function remains robust at scale and that our proposed scaling strategy effectively translates the benefits of HDS to large-scale language model pre-training.

\section{Conclusion}
We introduced the Holistic Data Scheduler (HDS), a novel online data mixing framework that reconceptualizes LLM pre-training as a multi-objective reinforcement learning problem in a continuous control space. By leveraging a sophisticated Soft Actor-Critic agent guided by a holistic reward function—synergistically integrating inter-domain influence to capture data synergies, scheduled lexical diversity to enforce an effective curriculum, and model stability to regularize updates—HDS consistently outperforms existing static and dynamic methods. Our extensive experiments on the Pythia suite demonstrate that HDS significantly accelerates training efficiency, matching the performance of static baselines with 57\% fewer iterations and achieving a remarkable 2.21x wall-clock speedup with negligible computational overhead. Beyond efficiency, HDS enhances final model quality, delivering a 13.6\% reduction in validation perplexity and substantial gains in downstream reasoning, including a 7.2\% improvement on MMLU (0-shot) and state-of-the-art results on commonsense tasks like COPA and PIQA. Furthermore, our validation on a 12B parameter model confirms the scalability of our approach, where HDS maintains a significant performance advantage over baselines. The success of HDS underscores the value of moving beyond single-objective optimization, establishing a more principled and data-centric paradigm for efficiently pre-training powerful and generalizable large language models. We believe HDS and the accompanying practical guidelines offer a valuable contribution to the community, paving the way for more cost-effective and robust LLM development.

\begin{acks}
Jing Ma's work was supported by the National Science Foundation of China (Grant Nos. 62272466 and U24A20233).
\end{acks}

%%
%% The next two lines define the bibliography style to be used, and
%% the bibliography file.
\bibliographystyle{acm-reference-format}
\bibliography{references}

%%
%% If your work has an appendix, this is the place to put it.
\appendix

\section{LLM Model Configuration}
Our language model is built upon a 16-layer Transformer architecture. It features a hidden dimension of 2048 and utilizes 16 attention heads, with all sequences processed at a fixed length of 1024 tokens. To encode positional information, we incorporate Rotary Positional Embeddings (RoPE)~\cite{su2024roformer}. For computational efficiency, the self-attention mechanism is implemented using FlashAttention~\cite{dao2022flashattention}, which significantly optimizes memory access patterns and reduces training time. Model optimization is performed using the Adam optimizer~\cite{adam2014method}. We employ a specific learning rate schedule, beginning with a linear warm-up over the first 833 training iterations from a base rate of 2.5e-5 to a peak of 2.5e-4. Subsequently, the learning rate follows a cosine decay schedule, eventually returning to the minimum rate of 2.5e-5. All text data is processed using the GPT-NeoX-20B tokenizer~\cite{black2022gpt}.

\section{Sensitivity Analysis of Reward Weights}
\label{sec:appendix_reward_weights}

\begin{table}[htbp]
  \centering
  \caption{Ablation of the three reward weight components (order: $w_{align}$, $w_{diversity}$, $w_{stability}$). Validation Perplexity is reported at 20,832 steps.}
  \label{tab:reward_weights_ablation}
  \begin{tabular}{lc}
    \toprule
    \textbf{Reward Weights} & \textbf{Validation Perplexity} \\
    \midrule
    $[1, 10, 10]$ & \textbf{12.73} \\
    $[0, 10, 10]$ & 13.45 \\
    $[1, 0, 10]$  & 12.89 \\
    $[1, 10, 0]$  & 12.79 \\
    $[2, 10, 10]$ & 12.81 \\
    $[1, 20, 10]$ & 12.78 \\
    $[1, 10, 20]$ & \textbf{12.73} \\
    \bottomrule
  \end{tabular}
\end{table}

To determine the optimal configuration for the multi-objective reward function, we investigated the sensitivity of the model's performance to variations in the scalar weights associated with each reward component: $w_{align}$, $w_{diversity}$, and $w_{stability}$. Table~\ref{tab:reward_weights_ablation} presents the validation perplexity at 20,832 steps for various weight combinations.

The baseline configuration of $[1, 10, 10]$ yields the optimal perplexity of 12.73. Consistent with our main ablation study, setting any single component's weight to 0 results in performance degradation, with the removal of the alignment weight ($w_{align}=0$) causing the most significant increase in perplexity to 13.45.

Furthermore, we explored the impact of amplifying the weights. Doubling $w_{align}$ to 2 or $w_{diversity}$ to 20 results in a slight regression in performance (12.81 and 12.78, respectively), suggesting that an over-emphasis on alignment or diversity may disrupt the learning balance. Interestingly, increasing the stability weight to 20 ($[1, 10, 20]$) maintains the optimal perplexity of 12.73. This indicates that while the stability reward is essential (as seen by the drop when $w_{stability}=0$), the model is relatively robust to higher magnitudes of stability regularization. Based on these findings, we selected $[1, 10, 10]$ as the default configuration to minimize hyperparameter complexity while ensuring peak performance.

\begin{figure*}[t]
    \centering
    \includegraphics[width=\textwidth]{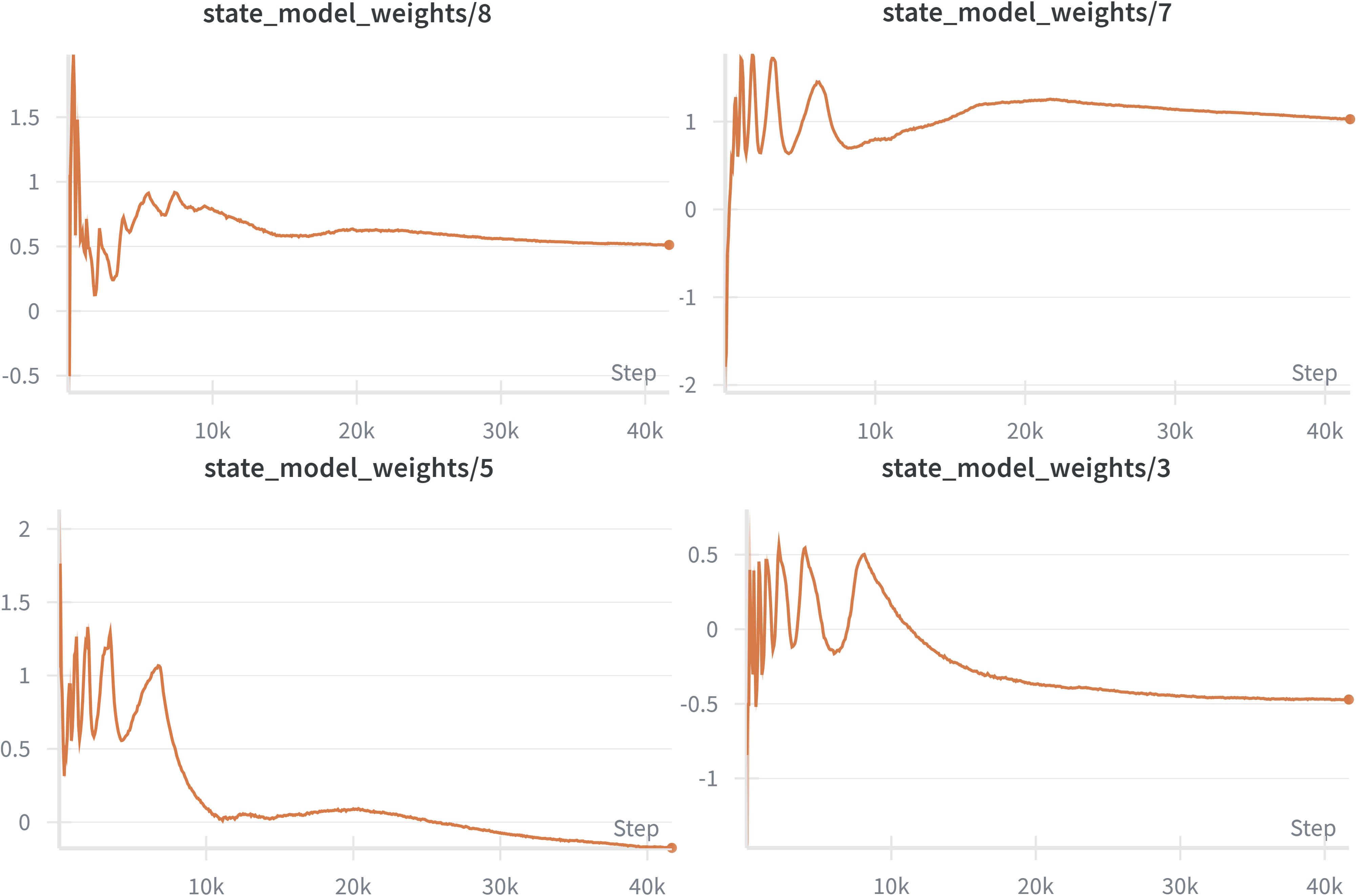}
    \caption{Evolution of the L2 norm for selected layer weights during the training process. Each subplot corresponds to a different set of layers used in the state representation. The consistent pattern of initial volatility followed by stabilization highlights the regularizing effect of the model stability reward ($r_{\text{stability}}$).}
    \label{fig:weight_norm_evolution}
    \Description{Evolution of the L2 norm for selected layer weights during the training process.}
\end{figure*}

\section{Hyperparameter Sensitivity Analysis}
\label{sec:appendix_c}
This section provides detailed results from our hyperparameter optimization experiments, specifically focusing on the layer selection strategies for the state representation and the alignment reward ($r_{\text{align}}$).

\subsection{Layer Selection Analysis}
We investigated how the choice of layers for constructing the model-centric components of our framework impacts overall performance. The results are summarized in Table~\ref{tab:state_layer_selection} and Table~\ref{tab:reward_layer_selection}.

Table~\ref{tab:state_layer_selection} details the ablation study on layer selection for the state vector's weight norm component. The results indicate a clear trend: incorporating information from a broader set of layers leads to a more comprehensive state representation and, consequently, better final perplexity. The optimal performance is achieved when all 24 layers are used. However, this approach incurs the highest computational cost. The layer sampling strategies, such as selecting all even or odd layers, provide a compelling alternative. They achieve a perplexity very close to the optimal result (e.g., 12.14 for even layers) while significantly reducing the computational overhead. This demonstrates that layer sampling is a highly efficient and practical method for constructing a rich yet manageable state representation, making it an excellent default choice for large-scale applications where computational resources are a primary concern.

\begin{table}[htbp]
  \centering
  \caption{Ablation study of selected layers for state representation. The final perplexity on The Pile test set is reported.}
  \label{tab:state_layer_selection}
  \begin{tabular}{lc}
    \toprule
    \textbf{Layer Selection Strategy} & \textbf{Perplexity} \\
    \midrule
    Standard HDS (Even Layers) & 12.15 \\
    All Layers (1-24) & \textbf{12.14} \\
    Odd Layers (1, 3, ..., 23) & 12.16 \\
    First Half (1-12) & 12.35 \\
    Second Half (13-24) & 12.28 \\
    \bottomrule
  \end{tabular}
\end{table}

\subsection{Alignment Reward}
Table~\ref{tab:reward_layer_selection} presents the results of our ablation study on the layer selection for calculating the alignment reward, $r_{\text{align}}$. The findings suggest that the choice of layers for this component has a more 
subtle, yet discernible, impact on performance.  This is intuitive, as deeper layers in a Transformer are known to capture more abstract and semantic information, which is more likely to represent the transferable knowledge that the alignment reward is designed to promote. Conversely, using earlier layers (e.g., blocks 2, 4, 6), which focus more on syntactic and local features, results in slightly worse performance. Nonetheless, the differences are not substantial, indicating that the HDS framework is relatively robust to the specific layer choice for this component. For optimal performance, however, selecting a few layers from the final third of the model is recommended.

\begin{table}[htbp]
  \centering
  \caption{Ablation study of selected layers for the alignment reward ($r_{\text{align}}$). The final perplexity on The Pile test set is reported.}
  \label{tab:reward_layer_selection}
  \begin{tabular}{lc}
    \toprule
    \textbf{Block Indexes} & \textbf{Perplexity} \\
    \midrule
    Standard HDS (12, 14, 16) & 12.15 \\
    2, 4, 6 & 12.20 \\
    8, 10, 12 & 12.19 \\
    14, 15, 16 & 12.15 \\
    \bottomrule
  \end{tabular}
\end{table}

\section{Analysis of Model Weight Norm Evolution}
To provide deeper insight into the training dynamics under HDS, we analyze the evolution of the L2 norm of the selected layer weights, a key component of our state vector. Figure~\ref{fig:weight_norm_evolution} illustrates these trajectories. A distinct and consistent pattern emerges across all monitored layers: an initial phase of high volatility is followed by a gradual convergence to a stable equilibrium. This behavior directly reflects the crucial role of the model stability reward, $r_{\text{stability}}$. The initial sharp fluctuations correspond to the early stages of training, where the model undergoes rapid and significant parameter updates as it begins to learn fundamental linguistic patterns. During this phase, the $r_{\text{stability}}$ component acts as a vital regularizer, penalizing overly drastic changes in the weight norms. By doing so, it discourages the SAC agent from selecting data mixtures that would lead to erratic or unstable updates. As training progresses, the agent learns to favor policies that promote smoother learning, causing the weight norms to stabilize. This demonstrates that the $r_{\text{stability}}$ reward effectively guides the training process towards a more stable and predictable trajectory, which is essential for preventing divergence and ensuring the model can robustly consolidate knowledge, ultimately contributing to the superior final performance of the HDS framework.

\end{document}